# An image processing analysis of skin textures

A. Sparavigna [1] and R. Marazzato [2,3]

[1] Physics Department, Politecnico di Torino, C.so Duca degli Abruzzi 24, Torino, Italy
[2] Department of Automation and Computer Science, Politecnico di Torino,
C.so Duca degli Abruzzi 24, Torino, Italy
[3] Faculty of Science and Technology, Free University of Bozen, V. Sernesi 1, Bolzano, Italy

**Abstract**
Colour and coarseness of skin are visually different. When image processing is involved in the skin analysis, it is important to quantitatively evaluate such differences using texture features. In this paper, we discuss a texture analysis and measurements based on a statistical approach to the pattern recognition. Grain size and anisotropy are evaluated with proper diagrams. The possibility to determine the presence of pattern defects is also discussed.

**Keywords**: 2D textures; skin texture; texture functions; defect localisation

## 1. Introduction

A quantitative characterisation of human skin textures is one of the tasks recently approached by image processing. This problem of texture analysis is twofold interesting. Besides the computational modelling of skin for realistic rendering in computer graphics [1], we must consider the possibility to apply the texture analysis to computer-assisted diagnosis in dermatology [2,3].

The skin texture is the appearance of the skin smooth surface. To the features of this texture, many factors are occurring, for instance diet and hydration, amount of collagen and hormones, and, of course, skin care. A gradual decline in skin is moreover superimposed by age. As skin ages, it becomes thinner and more easily damaged, with the appearance of wrinkles. The deterioration is also accompanied by a darkening of skin colour for an over-absorption of the natural colouring pigment, melanin, by the top most cell layer in skin. The skin texture also depends on its body location. In the case of image processing, we have to consider the fact that texture appearance is changing with image recording parameters, that is camera, illumination and direction of view, a problem common to any real surface.

The task to have a quantitative evaluation of the skin features is quite complex, as in all the cases where image analysis must be applied to surfaces with irregular non-periodic patterns. In the digital image processing, several methods have been developed to classify images and define statistical distances among them, with the aim to decide whether, in a set of many images, there exist some which are close to any arbitrary image previously encountered. The texture discrimination can be obtained by choosing a set of attributes, the texture features, which account for the spatial organisation of the image [4-7].

For skin textures, approaches based on wavelets [8], adaptive segmentation [9] and genetic image analysis [10] have been proposed. References [11] and [12] have faced the skin topography characterisation by processing the skin profile obtained with a capacitance device, to investigate the ageing of skin. Here, we propose to apply at skin characterisation, the image processing procedure previously used to investigate texture transitions in nematic liquid crystals [13-14]. This processing is suitable for images with smooth, scarcely regular textures, as those observed in the microscopy investigation of certain nematic liquid crystal cells. The processing is based on a coherence length analysis, described in detail in the following section.

**2. Image analysis**

To each pixel at the arbitrary point $P(x,y)$ in the image frame we associate a grey tone $b$ ranging from 0 to 255: $b(x,y)$ is then a 2-dimensional function representative of the image intensity (brightness) distribution. Starting from function $b(x,y)$, which gives the pixel grey-tone, the following calculation can be performed. First of all, the mean intensity of the pixel tones is determined:

$$M_o = \frac{1}{l_x l_y} \int_0^{l_x} \int_0^{l_y} b(x,y) \, dx \, dy \qquad (1)$$

where $l_x, l_y$ are the $x$- and $y$- rectangular range of the image frame. More generally, the $k$-rank statistical moments of the image are defined in the following way:

$$M_k = \frac{1}{l_x l_y} \int_0^{l_x} \int_0^{l_y} [b(x,y) - M_o]^k \, dx \, dy \qquad (2)$$

With this kind of characterisation we are then able to define the average values of the moments for the whole image frame. The distribution of pixel tones is then given according to these moments. The tone dispersion turns out to be evaluate by moment with $k=2$.

All integrals can be calculated on the whole image or on a window. In the case of windowing the image, moments $M_o$ and $M_k$ allow to find position and shape of objects, because the distribution can change for each specific window. In images where, at a first glance no particular objects are present, we can use the same values of the moments $M_o$ and $M_k$ defined by equations (1) and (2) for the whole image, supposing the image to be characterised by a single intensity distribution. However, to decide if an image exhibits irregular domains or localised defects, it is useful to estimate the quality ratio between the intensity standard deviation and the average intensity value, taken over the entire image frame, within a fixed acceptance limit, say, for instance of 50%.

Let us stress the fact that a domain can be described by an intensity distribution, which can essentially differ from that of other domains or from the background distribution. In this case, it is misleading to start from the point of view that only one distribution is enough to describe the whole image frame. On the contrary, it is necessary in this case to share the image in a lattice of windows, where inside each window the quality ratio is lower than the acceptance limit. The choice of the limit value does not affect the sensitivity of the method, as the choice connected with subdividing a statistical variable range of classes does not also affect position and dispersion indices of the sample.

However, it is necessary to check whether the hypothesis of homogeneity is really verified and whether preferred directions are displayed in the image frame (hypothesis of isotropy). We introduced a typical length characterising the texture size, which was very useful for characterisation of smectic and nematic phases [13,14].

Instead of measuring the homogeneity, by evaluating the histogram's entropy of intensity difference versus distance from a point of the image frame (see for instance [15]), or by calculating the spatial organisation by means of `run-length statistics' [16,17], we compute a set of coherence lengths defined in the following way. Starting from an arbitrary point $P(x,y)$ of the figure $b(x,y)$, along several radial directions, we calculate the values of $M_o^i(x,y)$ and $M_k^i(x,y)$ moments, that is:

$$M_o^i(x,y) = \frac{1}{l_{o,i}} \int_0^{l_{o,i}} b(x + r\sin\theta_i, y + r\cos\theta_i) dr \tag{3}$$

$$M_k^i(x,y) = \frac{1}{l_{o,i}} \int_0^{l_{o,i}} \left[ b(x + r\sin\theta_i, y + r\cos\theta_i) - M_o^i(x,y) \right]^k dr \tag{4}$$

where index $i$ is ranging over the radial directions, $r$ is the radial distance from $P$, and $\theta_i$ is the angle formed by $i$-direction with the $y$-axis (see Fig.1 for the frame of reference). The lengths $l_{o,i}$ and $l_{k,i}$ are the radial distances (from $P$) at which the values of the moments $M_o^i(x,y)$ and $M_k^i(x,y)$ on the chosen direction saturate, within a threshold level $t$, to the image average moments $M_o$ and $M_k$. This is the way to define the local "coherence lengths"[1] $l_{o,i}(x,y)$ and $l_{k,i}(x,y)$ of point $P$ in the image frame. The choice of threshold value $t$ depends on the problem under study.

In the calculation of the functions $l_{o,i}(x,y)$ and $l_{k,i}(x,y)$, the pixels near the image frame boundary are not involved, because in this case it would not be possible to estimate the coherence lengths in all the directions (boundary effect). On the contrary, in standard image processing techniques [18], periodicity of the image, originally present or artificially introduced by replication of the frame, is used to overcome the boundary problem. Let us stress the fact that moments $M_o^i(x,y)$ and $M_k^i(x,y)$ are not calculated on a window in the image frame, but on specific directions: therefore the method is different from the standard statistical approach, allowing to take into account, in a natural way, the anisotropy in the problem of texture recognition. In our analysis, we will use the 32 directions of Fig.1.

Actually, we can look for anomalous behaviours of vectors $l_{o,i}(x,y)$ or $l_{k,i}(x,y)$ as signals of the presence of a defect at the position in the image frame corresponding to given point $P(x,y)$. To discuss what can be properly considered as an anomalous behaviour of the coherence lengths, let us introduce the following average values of coherence lengths, averaged over the complete frame, for each specific $i$-direction:

$$L_{o,i} = \frac{1}{l_x l_y} \int_0^{l_x} \int_0^{l_y} l_{o,i}(x,y) dx dy \tag{5}$$

$$L_{k,i} = \frac{1}{l_x l_y} \int_0^{l_x} \int_0^{l_y} l_{k,i}(x,y) dx dy \tag{6}$$

---

[1] Coherence length is a well-known concept in optics and condensed matter physics. In optics, it is the propagation distance from a coherent source to a point where an electromagnetic wave maintains a specified degree of coherence. In condensed matter physics, it is the distance over which order is maintained. As an example, we can tell that we have coherence in a state of the matter when we have a long-range atomic or molecular order. Coherence lengths are significantly larger than molecular size. Normally, coherence length scales the size of ordered domains in material where long-range ordering occurs, as in liquid crystals for example. The term coherence length is also used for the scale characterising the profile of average molecular orientation in the distorted transition layers, formed at a solid/liquid-crystal interface when an electric or magnetic external field is applied.

If the image frame were strictly homogeneous, such averaged lengths should coincide with the actual local lengths measured for all image points. On the other hand, if the image frame were completely inhomogeneous, the local lengths would be very dispersed around their averages. The same occurs when the image frame is shared in windows, each of them characterised by a different intensity distribution. It is acceptable to average the coherence length over the whole image frame if the image can be considered as characterised by one distribution only, within a reasonable dispersion. The lengths $L_{o,i}$ represent the distance from a generic point $P(x,y)$ along the *i*-direction, at which the average value of the image intensity is practically reached: this means that the distance is dependent on the threshold level.

In Figure 2, the average values $L_{o,i}$ for two images of snake skins from the Brodatz album are reported. The result is a diagram showing $L_{o,i}$ in the 32 directions of Fig.1. We can define this diagram has the "coherence length diagram". In fact, the figure shows two diagrams obtained by fixing two different threshold values. To obtain the inner diagram we use a threshold corresponding to the 50% of ratio $\sqrt{M_2}/M_o$. The outer diagram is obtained with the 20% of the same ratio. The diagrams reveal preferential directions in the image texture, that is the anisotropy of the texture.

In this paper, we consider just $L_{o,i}$, because this is giving the most visually appreciable results. The diagram of $L_{o,i}$ lengths represents the smallest area about the generic point $P(x,y)$, on which area, when evaluating the average value of pixel intensity, we obtain the value $M_o$, within a fixed threshold level. The diagram represents the boundary of a unit area of the image, which contains the typical features of the whole image, as we can easily see comparing diagrams with the snake scales. In fact, the unit area is behaving as the primitive unit cell in crystal lattices (see Ref.19 for a description of unit cells in lattices). We can also consider this area, or the coherence length diagram, as a measure of the grain size and then evaluate the coarseness of the image texture.

A defect in the image texture could be considered any object in the image frame with a different shape for instance, different from the shape of the unit cell, or a different cell average colour tone, and so on.

**3. Analysis of skin textures and discussion.**
The case of snake skin gives diagrams immediately recalling the properties of the unit cells in crystal lattices. This is because the texture is quite geometric. The features of human skin are of course different, but as we observe from its use in microscopy investigation of liquid crystals [13,14], it is in the case of almost homogenous images, where Fourier analysis is scarcely active, that coherence lengths are useful.

We analysed with the coherence length ($L_{o,i}$) diagrams images of human-like textures and the result in shown in Fig.3, in the middle of the figure. In the lower part of the figure, it is shown a leather texture and the corresponding analysis. The inner and outer curves have thresholds as those used to obtain diagrams of Fig.2. The shape of the two diagrams does not substantially change. The area is changing: this is due to the fact that, to be fulfilled, a lower threshold requires a wider area.

On the right part of the figure, we see the detection of defects: the points marked in red are considered as "defects", whereas the pixels in green are the normal ones. It is not a segmentation procedure at the origin of the red-green maps, but a criterion involving the local behaviour of the coherence lengths $l_{o,i}(x,y)$. In the case that we are discussing, that is the case of an almost homogeneous image frame, we reasonably assumed that point $P(x,y)$ does not belong to a defect, if the average local value, defined as:

$$l_o(x,y) = \frac{1}{32}\sum_{i=1}^{32} l_{o,i}(x,y) \qquad (7)$$

is included between two extremal lengths $L_{o,min}$ and $L_{o,max}$, where $L_{o,min}$ is the minimum of the 32 average values of equation (5) and $L_{o,max}$ is the maximum. Obviously, for an almost isotropic image frame, the two extremal values $L_{o,min}$ and $L_{o,max}$ are close together.

At points marked in red, which are the defects, the average local coherence length $l_o(x,y)$ is not comprised in interval $[L_{o,min}, L_{o,max}]$. Instead, points not belonging to defects are marked in green. We use red and green half-tones to see also the original texture in this defect map.

With commercial software, common procedures to identify defects are based on thresholds of grey tone [15]: this means a procedure checking whether the pixel intensity is, within a fixed tolerance, coincident or not with a specific chosen grey tone. The processing is known as "image segmentation by thresholding" and produces images segmented in two or more regions according to the used thresholds. This technique is not investigating the neighbourhood of the pixel and then it is impossible to ascertain if it is truly belonging to a defect or not. With the analysis here discussed, the detection of defects is a comparison of the local coherence lengths $l_{o,i}(x,y)$, that is of a local unit cell, with the coherence length $L_{o,i}$ diagram, the global unit cell, which is shown in the middle of Figure 3. Then, we are comparing the behaviour of a pixel neighbourhood with the average behaviour of all the pixel neighbourhoods.

In the case of human skin, a defect could be a region with a paler or darker colour or a region with wrinkles, for instance. In the upper part of the Figure 3, we see an almost regular texture, with a darker region. The coherence length diagrams are telling that the texture is isotropic, and, actually we have not wrinkles. The defect map obtained as previously explained, puts in evidence the darker regions.

Figure 3 shows that a texture with wrinkles has a slightly anisotropic unit cell, as the coherence length diagrams report (see the middle of the figure). Neglecting this anisotropy, the procedure of defect detection sees the regions with different intensity from the background. A procedure can be developed to compare the true shape of the local coherence area, with the global one, but this is the subject of further studies. The leather surface has no wrinkles and then behaves as the upper image in the same figure. We could consider this defect detection procedure as a good means for fault detection in leather industry.

It is possible to obtain the coherence length diagram of the maps obtained from the capacitance system too (see Ref.11). In the upper part of Figure 4, we can see a map as it is from the capacitance system. In the lower part, we see the same image after a normalization of the image contrast: the capacitance image is sensitive to different hydration and to the presence of sweat, which give darker regions, and then a renormalization is required.

Note that the coherence length diagrams are different, because the pixel tone distributions are different. This difference is put in evidence by the procedure of defect detection that we have previously used for the textures shown in Fig.3. In the right-upper part of Fig.4, we see the red areas concentrated where the normalization procedure must much more strongly act in changing the pixel tone distribution, to give the renormalized image. Accordingly, the right-lower image, which is the defect map of renomalized image, shows a small amount of defects.

The aim of Ref.11 was the development of a device to characterise the skin topography to measure the skin profile and the presence of wrinkles. The renormalization of images is then necessary for the segmentation of skin topography, to correlate it with skin ageing. In the case of processing images recorded by cameras, where illumination and direction of view are important, a similar normalization procedure can be quite useful too.

Our image analysis of skin texture is based on the evaluation of the global grey tone distribution of whole image frame and then on the coherence length diagrams which are able to show anisotropy of the texture. The diagrams are also able to estimate texture features, such as anisotropy and coarseness: these diagrams can then adequately describe the presence of wrinkles. According to the chosen statistic parameters of grey tone distribution, several procedures to defect detection can be proposed. We followed a simple procedure able to identify local differences from the background, but more sophisticated procedures, suggested by clinical experience can be easily approached.

**Acknowledgements**. Many thanks are due to A. Bevilacqua and A. Gherardi for images used in Figure 4.

**References**
[1] Cula O.G., Dana K.J., Murphy F.P., Rao B.K., 2005, Int. J. Computer Vision, 62, 97.
[2] Jun Liu, Bowyer K., Goldgof D., Sarkar S., 1997, Information, Communications and Signal Processing, ICICS '97, Singapore, 9-12 September 1997.
[3] Pope T.W., Williams W.L., Wilkinson S.B., Gordon M.A., 1996, Computer and Biomedical Res., 29, 429
[4] Haralick R.M. 1979, Proc. IEEE, 67, 786.
[5] Weszka J.S., Der C.R., Rosenfeld A., 1976, IEEE Trans. Syst. Man. Cybern. 6, 369.
[6] Azencott R., Wang J., Younes L., 1997, IEEE Trans, Pattern Anal. Mach. Intell. 19, 148.
[7] Haralick R.M., Shanmugan K., Dinstein I., 1973, IEEE Trans. Syst. Man. Cybern. 3, 610.
[8] Doi M., Tominaga S., 2006, IEEE Southwest Symposium on Image Analysis and Interpretation, Pages:193-197.
[9] Son Lam Phung, Chai D., Bouzerdoum A., 2003, Proceedings IEEE lnternational Conference on Acoustics, Speech, & Signal Processing. April 6-10, 2003, Hong Kong. Pages:III173-176.
[10] Nishioka M., Fukumi M., Akamatsu N., Mitsukura Y., 2004, Proceedings Int. Symposium on Intelligent Signal Processing and Communication Systems, ISPACS 2004, 18-19 Nov. 2004 Pages:787-791.
[11] Bevilacqua A., Gherardi A., 2004, Proceedings of the 17$^{th}$ International Conference on Pattern Recognition (ICPR'04), Pages:1-4.
[12] Bevilacqua A., Gherardi A. , Guerrieri R. 2005, Proceedings of the Seventh IEEE Workshop on Applications of Computer Vision (WACV/MOTION'05), Pages:1-6.
[13] Montrucchio B., Sparavigna A., Strigazzi A., 1998, Liquid Crystals 24, 841.
[14] Sparavigna A., Mello A., Montrucchio B., 2000, Recent Res. Devel. Patterns Rec., 1, 29.
[15] Pitas, I. 1993, Digital Image Processing (Berlin: Springer Verlag).
[16] Levine M.D., 1985, Vision inMan and Machine (Mc-Graw Hill).
[17] Galloway M.M., 1975, Comp. Graph. Image Process, 4, 172.
[18] Jähne B., 1993, Digital Image Processing (Berlin, Springer Verlag).
[19] Ashcroft N.W., Mermin N.D., 1976, Solid State Physics (Saunders College Publishing)

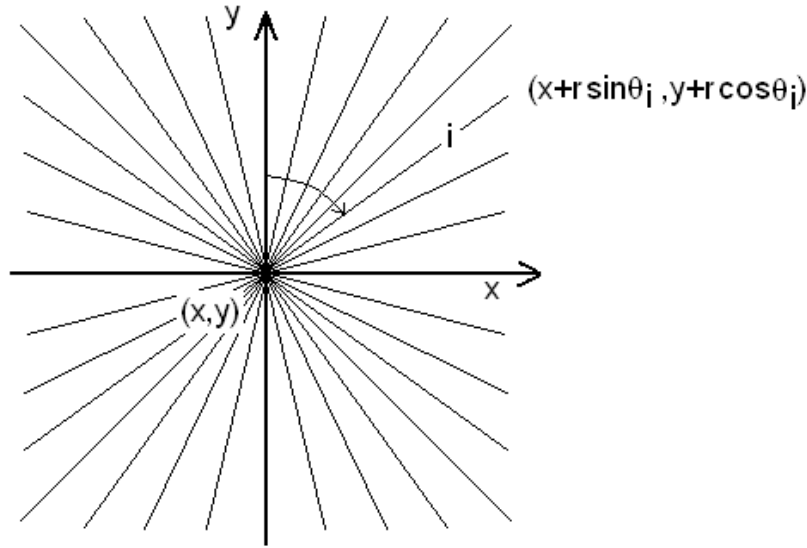

Figure 1. Reference system used in the calculation of average values along *i*-direction. In the figure we show the 32 directions used in our evaluations of coherence length diagrams.

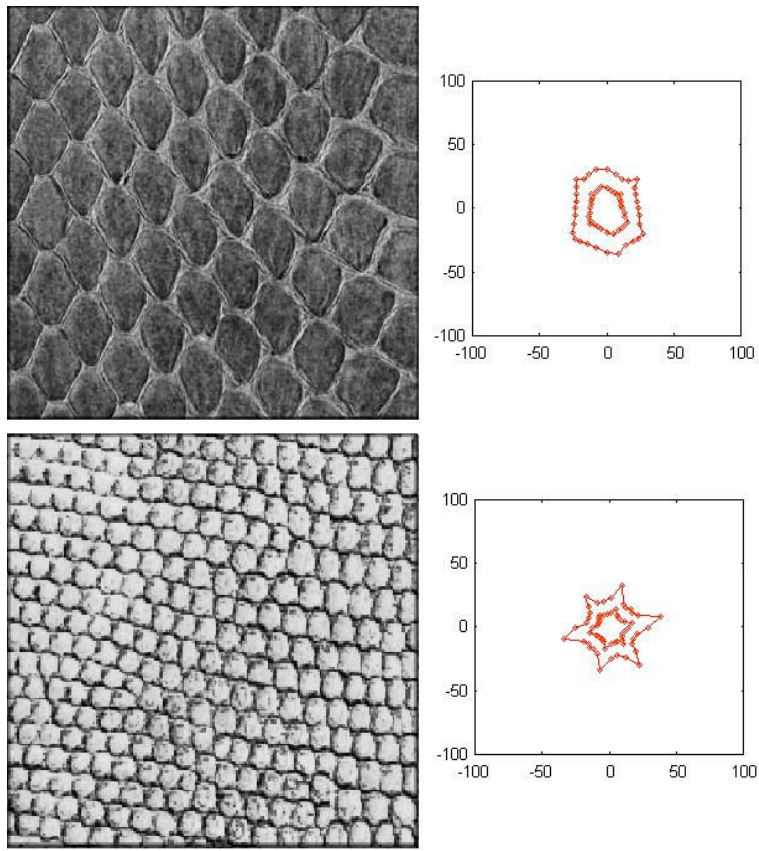

Figure 2. Coherence length diagrams (on the right) of two snake skins from the Brodatz album. The inner curve corresponds to a threshold of $0.5\sqrt{M_2}/M_o$, the outer to $0.2\sqrt{M_2}/M_o$. The numbers on axes correspond to pixel numbers. Note that the graph is able to indicate the texture anisotropy. For the chosen threshold, diagrams are the boundary of the smallest area having the same features characterising the whole image frame.

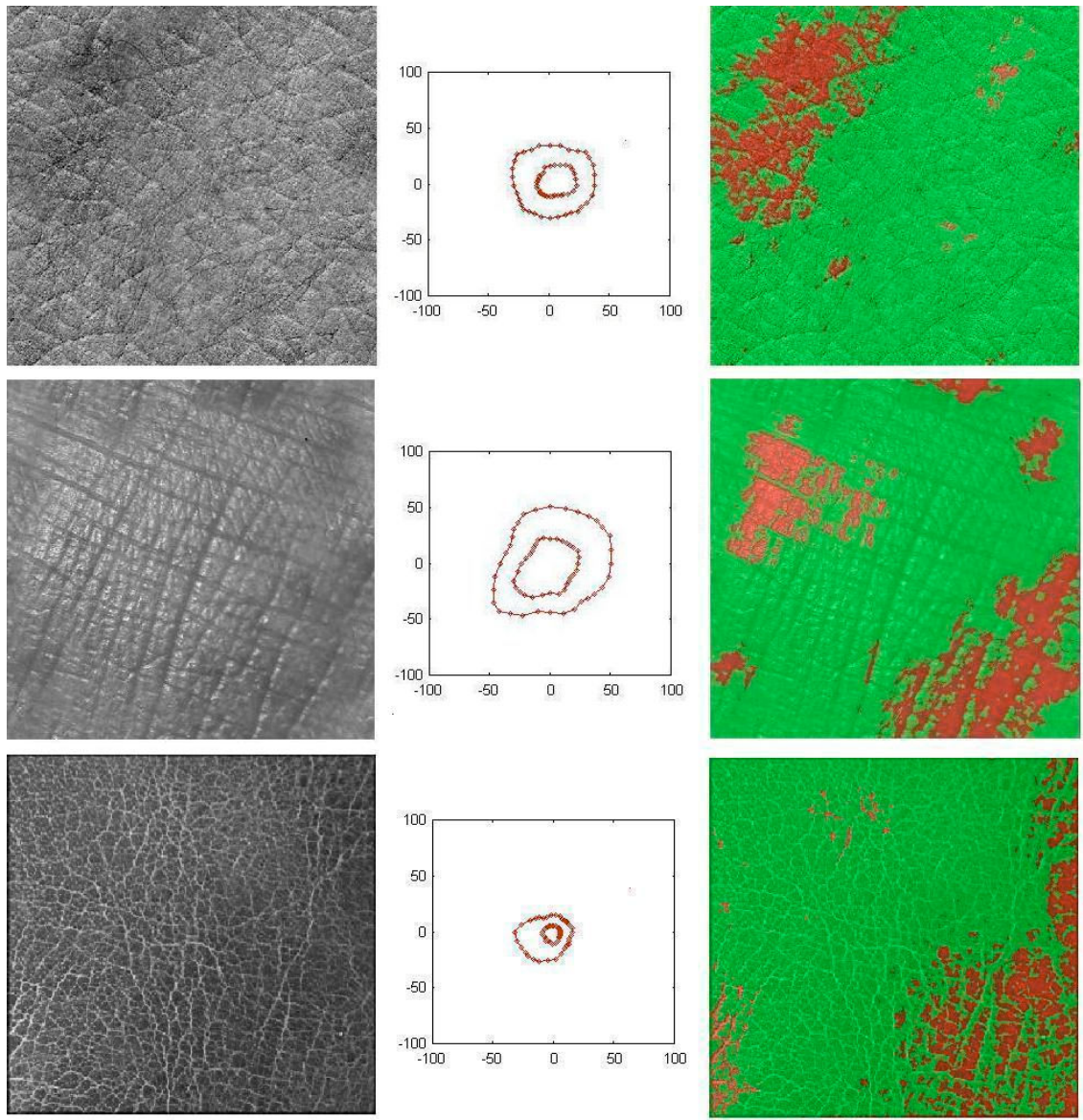

Figure 3: In the middle of the figure, we can see coherence length diagrams of two human-like skin textures and of a leather texture (lower image). The inner and outer curves have thresholds as in Fig.2. On the right part of the figure, we see the detection of "defects". Points marked in red are considered as "defects" (see text for explanation). Let us note that this is not a segmentation threshold procedure, as it is possible to obtain with commercial programs.

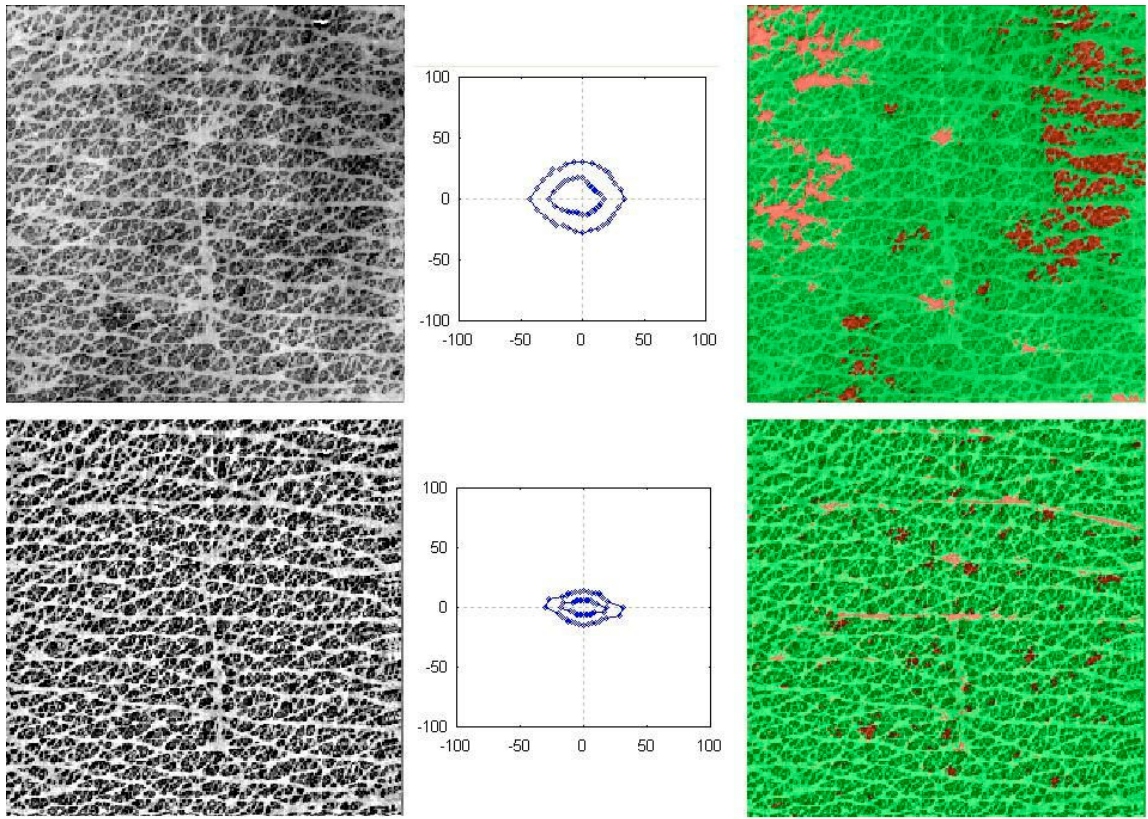

Figure 4. It is possible to use the coherence length diagram evaluation, on maps obtained by means of a capacitance system (from Ref.11). The diagrams have threshold values as in Figs. 2 and 3. The image at the left-upper part shows the map as it is from the capacitance system. In the lower part of the figure, we see the same image after a normalization of image contrast. Note that the diagrams are different, because the pixel tone distributions are different. This difference is enhanced by the procedure of defect detection (the procedure is the same as for Fig.3). In the upper part, we see the red areas concentrated where the renormalization procedure must act in changing the pixel tone distribution. In the lower image, the number of defects is strongly reduced.